\documentclass[table, 11pt]{article}
\usepackage[table, svgnames, dvipsnames]{xcolor} % Gộp xcolor
\usepackage{tikz}
\usepackage[justification=centering]{caption}
\usepackage[most]{tcolorbox}
\colorlet{promptbg}{green!15}
\usepackage[colorlinks=true, citecolor=blue, linkcolor=blue, urlcolor=blue]{hyperref}
\usepackage[final]{acl}
\usepackage{amsfonts} % For \mathbb{R}
\usepackage{booktabs} % For better table rules
\usepackage{multirow} % If needed for table structure, though not strictly for this one
\usepackage{graphicx} % ở phần preamble nếu chưa có
\usepackage{amsfonts} % For \mathbb{R}
\usepackage{booktabs} % For professional tables
\usepackage{multirow} % For multi-row cells in tables
\usepackage{graphicx} % Required for includegraphics (though not used in text here)
\usepackage{xcolor} % For coloring text
\usepackage{bold-extra} % For bold text if needed, or just use \textbf
\usepackage{times}
\usepackage{latexsym}
\usepackage{enumitem}
\usepackage[T1]{fontenc}
\usepackage[utf8]{inputenc}
\usepackage{microtype}
\usepackage{inconsolata}
\usepackage{graphicx}
\usepackage{amsmath}
\usepackage{amsfonts}
\usepackage{bm}
\usepackage{multirow}
\usepackage{booktabs}
\usepackage{xspace}
\usepackage{adjustbox}
\usepackage{float}
\usepackage{stfloats}
\usepackage{natbib} % Thêm natbib để hỗ trợ \citep, \citet

\usepackage{colortbl}
\usepackage{subfigure}

\usepackage[linesnumbered,ruled,lined]{algorithm2e}

\definecolor{lightgray}{gray}{0.80}
\definecolor{darkgreen}{rgb}{0.0, 0.5, 0.0}
\newcolumntype{g}{>{\columncolor{lightgray}}c}

\title{TALAS: Teacher-Anchored Layer Alignment with Adaptive Sharpness-Aware Minimization for Embedding Distillation}

\author{
  \textbf{Quoc Phong Dao\textsuperscript{1}\footnotemark[1]},
    \textbf{Hoang Son Nguyen\textsuperscript{1}\footnotemark[1]},
  \textbf{Pham Khanh Chi\textsuperscript{1}\footnotemark[1]}, \\
    \textbf{Linh Ngo Van\textsuperscript{1,\dag}},
  \textbf{Diep Thi-Ngoc Nguyen\textsuperscript{2}},
  \textbf{Thien Huu Nguyen\textsuperscript{3}},
  \textbf{Trung Le\textsuperscript{4}}
  \bigskip \\
\textsuperscript{1}Hanoi University of Science and Technology, \\
\textsuperscript{2}VNU University of Engineering and Technology,
\textsuperscript{3}University of Oregon,
\textsuperscript{4}Monash University
}

\begin{document}

\maketitle
\renewcommand{\thefootnote}{\fnsymbol{footnote}}
\footnotetext[1]{Equal contribution}
\footnotetext[2]{Corresponding author: \href{mailto:email@domain}{ linhnv@soict.hust.edu.vn}}
\renewcommand*{\thefootnote}{\arabic{footnote}}
\begin{abstract}
    Knowledge Distillation (KD) has established itself as a pivotal technique for compressing large pre-trained language models. However, existing methods that force a student to strictly mimic the teacher’s sentence embeddings or internal features often incur prohibitive computational costs and yield suboptimal performance due to the inherent capacity gap.
To address these challenges, we propose \textbf{TALAS} (\textbf{T}eacher-\textbf{A}nchored \textbf{L}ayer \textbf{A}lignment with \textbf{S}harpness-aware minimization), a unified framework that synergizes hierarchical (multi-layer) alignment with robust optimization.
First, we introduce a Teacher-Anchored mechanism that selectively distills final sentence embeddings only into the student's upper layers, thereby reducing overhead while respecting capacity constraints. Second, we bridge the semantic gap in lower layers via Layer-Aligned Self-Distillation, which propagates knowledge top-down using internal geometric relational constraints in the embedding space. Finally, to prevent the student from memorizing point-wise teacher noise, we integrate Adaptive Sharpness-Aware Minimization (ASAM) into the training objective, guiding the model towards flat minima for enhanced generalization.
Empirical results on standard sentence embedding benchmarks demonstrate that TALAS consistently outperforms strong distillation baselines while achieving superior training efficiency in terms of computational cost and memory footprint.
\end{abstract}
\section{Introduction}\label{sec:intro}
% thêm related work vào đây nữa

With the rapid advancement of natural language processing, text embedding models have become a core component of Retrieval-Augmented Generation (RAG) systems, enabling effective semantic retrieval to supply relevant context for large language models \citep{gao2024retrievalaugmentedgenerationlargelanguage,zhao2025surveylargelanguagemodels}. They play a critical role in downstream tasks such as information retrieval, question answering, and text classification by facilitating meaning-based comparison and representation of textual data \citep{ramesh-kashyap-etal-2024-comprehensive}. Despite their strong performance on benchmarks such as MTEB \citep{muennighoff-etal-2023-mteb} state-of-the-art text embedding models are often characterized by large parameter counts and high-dimensional representations, which substantially increase computational and deployment costs. 

% \footnote{\url{https://huggingface.co/spaces/mteb/leaderboard}}. 

In parallel, knowledge distillation \citep{hinton2015distillingknowledgeneuralnetwork} has emerged as a fundamental technique for compressing large neural models, with recent advances primarily focusing on large language models. However, distillation methods specifically tailored to embedding models - despite their central role in many downstream tasks \citep{ramesh-kashyap-etal-2024-comprehensive} - remain comparatively less explored and less focused. A key challenge in embedding model distillation lies in the trade-off between effectiveness and efficiency. Approaches that exploit internal representations such as hidden states or attention patterns provide rich supervision and have been shown to achieve strong performance in knowledge transfer settings \citep{  sun2019patientknowledgedistillationbert, jiao2020tinybertdistillingbertnatural, wang2020minilmdeepselfattentiondistillation, passban2020alpkdattentionbasedlayerprojection, zhang-etal-2024-intermediate, dasgupta2025improving, zhao-etal-2025-analysis}. However, when the teacher is a large LLM-based embedding model, leveraging such intermediate signals typically incurs substantial computational and memory overhead, which limits their practicality in large-scale or resource-constrained scenarios. In contrast, methods that rely solely on the teacher’s output representations offer a significantly more efficient training pipeline and are particularly attractive when the teacher is only accessible through inference \citep{hinton2015distillingknowledgeneuralnetwork, ko2024distillmstreamlineddistillationlarge, anshumann-etal-2025-sparse, vu2026dwa}. Nevertheless, compressing rich semantic information into a single output embedding, combined with the capacity gap between teacher and student models, can lead to the student failing to capture all of the important nuances present in the teacher’s distribution; instead, the student may focus on more dominant or easier aspects of the teacher’s outputs, as discussed in prior work on distillation objectives for generative models \citep{gu2024minillmknowledgedistillationlarge}.

A recurring obstacle in knowledge distillation is the capacity gap between a high-capacity teacher and a compact student: when the student lacks sufficient expressiveness, a stronger teacher does not necessarily produce a better distilled model and can even degrade performance \citep{cho2019efficacyknowledgedistillation}. This issue becomes particularly salient for embedding distillation with LLM-based teachers, where directly forcing the student to mimic the teacher across the network can be unstable - especially in shallow layers that primarily encode low-level linguistic features - because aligning them to highly abstract teacher representations may disrupt feature extraction and hinder optimization. Prior work has proposed using intermediate assistant models to help bridge the gap between large teachers and compact students and to stabilize the distillation process by gradually reducing representational differences through one or more auxiliary networks \citep{mirzadeh2019improvedknowledgedistillationteacher, son2021denselyguidedknowledgedistillation, zhou-ai-2024-teaching}.
 Inspired by this principle, we adopt a progressive bridging strategy: we restrict teacher-anchored supervision to the student’s upper layers while using the student’s own higher layers as a lightweight “assistant” signal to guide lower layers through sequential, top-down alignment. This design directly targets the capacity-gap bottleneck while remaining compatible with resource-constrained settings where accessing or storing teacher internals is often impractical. Moreover, prior studies suggest that sharp minima in the loss landscape are closely associated with larger generalization gaps, motivating us to use Adaptive Sharpness-Aware Minimization (ASAM) \citep{kwon2021asamadaptivesharpnessawareminimization} of loss-surface sharpness as a diagnostic and optimization signal. In the context of embedding distillation, where generalization across domains and tasks is crucial, controlling sharpness provides a promising direction for balancing efficiency and performance. 
Our main contributions include:
\begin{itemize}
    \item We propose \textbf{TALAS} (\textbf{T}eacher-\textbf{A}nchored \textbf{L}ayer \textbf{A}lignment with \textbf{S}harpness-aware minimization) - a resource-efficient unsupervised knowledge distillation framework for embedding models that aligns multiple student layers with teacher output embeddings, eliminating the need for teacher inference during training.
    \item We introduce a layer-aligned self-distillation mechanism that propagates relational structure across student layers without relying on teacher hidden states. Additionally, we incorporate adaptive sharpness-aware optimization to mitigate the capacity gap between large teacher models and compact students, improving generalization.
    \item We empirically demonstrate effective unsupervised distillation from large LLM-based embedding models to BERT-based students using only cached teacher embeddings.
\end{itemize}

\section{Background}
This section introduces the background concepts underlying our approach. 
We first review the fundamentals of knowledge distillation, followed by Adaptive Sharpness-Aware Minimization (ASAM). An extended discussion of related work is in appendix \ref{app:erw}.

\paragraph{Knowledge Distillation.}
Knowledge Distillation aims to transfer knowledge from a teacher model $T$ to a student model $S$ by encouraging the student to approximate the teacher’s behavior under a chosen supervision signal. In general, given an input $x$, the teacher produces an output $z_T(x)$ and the student produces a corresponding output $z_S(x)$, where $z(\cdot)$ may represent logits, hidden representations, or embeddings depending on the task. The student is trained by minimizing a distillation loss of the form
\begin{equation}
\mathcal{L}_{\text{KD}} = \mathcal{D}\bigl(z_S(x), z_T(x)\bigr),
\end{equation}
where $\mathcal{D}(\cdot,\cdot)$ denotes a discrepancy measure such as cross-entropy, mean squared error, or a similarity-based distance.

\paragraph{Adaptive Sharpness-Aware Minimization.}
Adaptive Sharpness-Aware Minimization (ASAM) is an optimization framework designed to improve generalization by encouraging convergence to flat minima. Instead of minimizing the empirical loss $\mathcal{L}(w)$ directly, ASAM solves a local min-max problem:
\begin{equation}
\min_{w} \; \max_{\|\mathbf{T}_w^{-1} \epsilon\|_2 \le \rho} \mathcal{L}(w + \epsilon),
\label{eq:asam_talas}
\end{equation}
where $w$ denotes model parameters, $\epsilon$ is an adversarial perturbation, $\rho$ controls the neighborhood size, and $\mathbf{T}_w$ is a parameter-dependent scaling matrix that normalizes perturbations according to parameter magnitude. By accounting for parameter scale, ASAM provides a more stable and adaptive variant of sharpness-aware optimization. This property is especially beneficial in knowledge distillation settings with strong supervision signals or large teacher-student capacity gaps.
\section{Methodology}
\begin{figure*}
    \centering
    \includegraphics[width=1.0\linewidth]{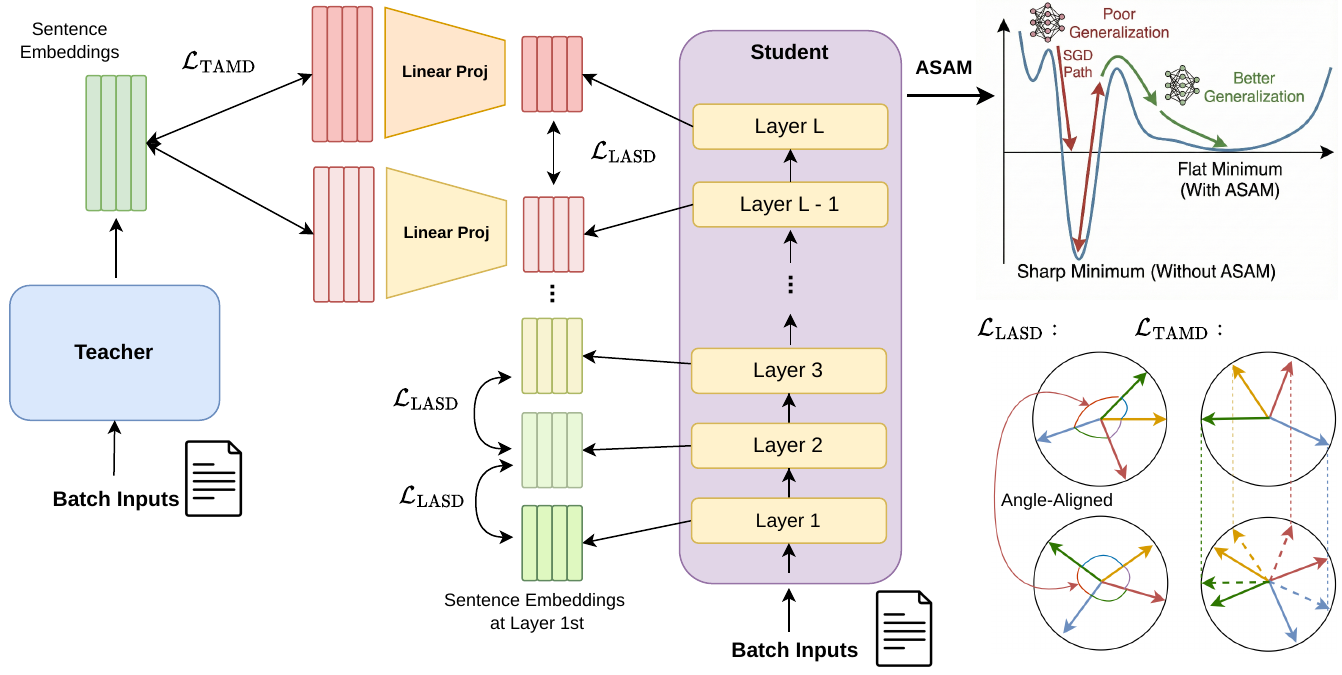}
    \caption{An illustration of the proposed TALAS framework. The training process synergizes two complementary distillation strategies: (1) \textbf{Teacher-Anchored Multi-Layer Distillation}($\mathcal{L}_{\text{TAMD}}$) aligns the student's upper layers with the teacher's stationary embedding via learnable linear projections; and (2) \textbf{Layer-Aligned Self-Distillation} ($\mathcal{L}_{\text{LASD}}$) enforces geometric structural consistency between adjacent student layers. The entire training is optimized with \textbf{Adaptive Sharpness-Aware Minimization (ASAM)} to seek flat minima for better generalization.}
    \label{fig:TALAS}
\end{figure*}

% In this section, we present TALAS, a unified framework for robust embedding distillation. To maximize knowledge transfer from the teacher while mitigating the inherent capacity gap, our method decomposes the training objective into three complementary components. First, Section \ref{sec: TAMD} introduces a Teacher-Anchored mechanism that aligns only the student’s upper layers with the teacher. Second, Section \ref{sec: LASD} details a Layer-Aligned Self-Distillation strategy to guide lower layers using internal representations. Finally, Section~\ref{sec: ASAM} describes the integration of Adaptive Sharpness-Aware Minimization (ASAM) to enhance generalization. An overview is depicted in Figure~\ref{fig:TALAS}.

In this section, we present \textbf{TALAS}, a unified framework for robust embedding distillation under teacher--student capacity mismatch. TALAS decomposes the training objective into three complementary components.
Section~\ref{sec: TAMD} introduces Teacher-Anchored Distillation, which defines \emph{where} semantic alignment with the teacher is applied.
Section~\ref{sec: LASD} presents Layer-Aligned Self-Distillation, which governs \emph{how} semantic knowledge is propagated across layers.
Finally, Section~\ref{sec: ASAM} integrates Adaptive Sharpness-Aware Minimization (ASAM) to enable robust optimization.
An overview is shown in Figure~\ref{fig:TALAS}.

\paragraph{Problem Setup.} Let $\mathcal{T}$ and $\mathcal{S}$ denote the teacher and student models, respectively. We define $\mathbf{e}^\mathcal{T}$ and $\mathbf{e}^\mathcal{S}$ as their corresponding sentence embeddings. For the student, let $\mathbf{e}^\mathcal{S}_l$ represent the pooled embedding at layer $l$. Depending on the backbone architecture, these representations are derived via specific pooling strategies, such as \texttt{[CLS]}~\cite{devlin2019bertpretrainingdeepbidirectional}, \texttt{[EOS]}~\cite{zhang2025qwen3embeddingadvancingtext}, or mean pooling~\cite{lee2024nv}.

\subsection{Teacher-Anchored Multi-Layer Embedding Distillation}
\label{sec: TAMD}

Unlike prior distillation methods that rely on complex internal signals such as attention maps and hidden states (e.g., EOFD~\cite{zhao2025analysis}, FDD~\cite{gong2025beyond}), our approach focuses exclusively on sentence embedding distillation, significantly reducing computational overhead. To compensate for the reduced supervision and maximize semantic transfer, we draw inspiration from the representation compression principle proposed in ESE~\cite{li2025ese}, which advocates for encoding salient semantics into lower network layers. Accordingly, we aim to guide both intermediate and final layers of the student to align with the teacher. However, acknowledging the inherent capacity mismatch, directly forcing alignment across all layers can be suboptimal. Therefore, we apply Teacher-Anchored distillation strictly to the upper layers, where the student possesses sufficient capacity to approximate the teacher’s embedding space.

Formally, we treat the teacher's final embedding as a stationary \textit{anchor} to supervise the student's representations across multiple upper layers. The teacher-anchored distillation objective is formulated as follows:

\begin{equation}
\begin{aligned}
\mathcal{L}_{\text{TAMD}}(l) &=  \frac{1}{N} \sum_{i=1}^{N}
\mathcal{D}_{\text{cos}} \left( \mathbf{e}_{l, i}^\mathcal{S} \mathbf{W}_l, \mathbf{e}_i^\mathcal{T} \right) \\
\mathcal{L}_{\text{TAMD}} &= \frac{1}{k+1} \sum_{l=L-k}^{L} \mathcal{L}_{\text{TAMD}}(l)
\end{aligned}
\label{eq:tamd_loss}
\end{equation}

\noindent where $L$ is the total number of student layers, $N$ is the batch size, $\mathcal{D}_{\text{cos}}(u, v) = 1 - \frac{u^\top v}{\|u\| \|v\|}$ denotes the cosine distance and $\mathbf{W}_l \in \mathbb{R}^{d_S \times d_T}$ is a learnable projection matrix at $l$-th layer.

In this formulation, $\mathbf{e}^\mathcal{T}$ serves as a shared semantic anchor that guides both intermediate and final representations. By anchoring multiple upper layers to the same teacher embedding, the student is encouraged to progressively align its semantic space with that of the teacher, rather than relying solely on supervision at the final layer. 
This design facilitates \textit{early semantic injection} during the forward pass, enabling intermediate layers to acquire high-level semantic information before it is fully consolidated at the output layer. Compared to single-layer distillation, such multi-layer anchoring stabilizes training and reduces representation drift across layers.

% Crucially, we restrict this teacher-anchored supervision to the top $k$ layers of the student (i.e., from $L-k$ to $L$). This selective alignment explicitly accounts for the \textit{representational disparity} between teacher and student models, as lower layers often lack the complexity to faithfully approximate the teacher’s final embedding space. By decoupling teacher supervision from lower layers, our approach avoids over-constraining early representations, providing a robust initialization for the layer-aligned self-distillation described next.

\subsection{Layer-Aligned Embedding Self-Distillation}
\label{sec: LASD}

While Section~\ref{sec: TAMD} anchors the upper layers to the teacher’s semantic space, extending this supervision to lower layers is non-trivial. Prior study on layer-wise distillation \citep{sun2019patientknowledgedistillationbert} have shown that directly forcing shallow student layers to match highly abstract teacher representations under large capacity gaps can destabilize training and degrade feature learning. Early layers primarily encode low-level linguistic patterns and directly forcing them to approximate the teacher’s highly abstract representations can distort their feature extraction behavior. At the same time, leaving the lower layers entirely unsupervised introduces a representational discontinuity across depth: the abrupt transition from unsupervised shallow layers to teacher-anchored upper layers yields structurally misaligned internal geometries, impeding the smooth propagation of semantic structure through the network.

To overcome this limitation, we propose a progressive self-distillation mechanism that propagates knowledge sequentially from top to bottom. Instead of mapping lower layers directly to the teacher, we utilize the student's own upper layer ($l+1$) as a dynamic ``guide'' for the immediate lower layer ($l$). This chain-based approach ensures a smooth semantic transition, allowing deep layers to maintain high-level abstractions while gradually adjusting shallow layers to the target embedding space without destroying their intrinsic structural properties.

We formulate this process as a geometric structural consistency between adjacent student layers problem using Relational Knowledge Distillation (RKD). Unlike point-wise distillation, RKD focuses on preserving the structural relations between data samples within a batch. 
Let $\mathbf{E}^\mathcal{S}_l \in \mathbb{R}^{N \times d}$ denote the batch of embeddings at layer $l$ of student model, where $N$ is the batch size. We first apply $L_2$ normalization to obtain $\tilde{\mathbf{E}}^\mathcal{S}_l$ and compute the relational matrix $\mathbf{R}^\mathcal{S}_l \in \mathbb{R}^{N \times N}$, which captures the pair-wise similarity structure:
\begin{equation}
    \mathbf{R}^\mathcal{S}_l = \tilde{\mathbf{E}}^\mathcal{S}_l (\tilde{\mathbf{E}}^\mathcal{S}_l)^\top,
\end{equation}
where each element $(\mathbf{R}^\mathcal{S}_l)_{ij}$ corresponds to the cosine similarity between the $i$-th and $j$-th samples, i.e., $\cos(\mathbf{e}_i, \mathbf{e}_j)$.

Here, $\mathbf{R}^\mathcal{S}_l$ represents the geometric topology of the embedding space at layer $l$. Our objective is to minimize the structural divergence between adjacent layers. The Layer-Aligned Self-Distillation loss is defined using the Frobenius norm of the difference between relation matrices:
\begin{equation}
    \mathcal{L}_{\text{LASD}} = \frac{1}{L-1} \sum_{l=1}^{L-1} \frac{1}{N^2} \left\| \mathbf{R}^\mathcal{S}_{l+1} - \mathbf{R}^\mathcal{S}_{l} \right\|_F^2,
    \label{eq:lasd_loss}
\end{equation}
where $N$ is the batch size and $\|\mathbf{A}\|_F^2 = \sum_{i,j} A_{ij}^2$ denotes the squared Frobenius norm, serving as a matrix-level Mean Squared Error proxy.

By optimizing Eq.~\ref{eq:lasd_loss}, the student model creates a seamless \textit{semantic bridge}, where the teacher's knowledge (anchored at the top) is progressively distilled down to the input level through structural alignment, bridging the capacity gap effectively.

% In summary, our framework employs a decoupled yet complementary strategy to distill knowledge. While \textbf{Teacher-Anchored Distillation} (Sec.~\ref{sec: TAMD}) imposes explicit high-level semantic guidance from the teacher (external supervision), \textbf{Layer-Aligned Self-Distillation} enforces structural consistency within the student's own hierarchy (internal supervision). By separating these two objectives, we effectively mitigate the risk of over-constraining the lower layers with direct teacher targets, allowing the student to construct a smooth semantic manifold that is both faithful to the teacher and internally coherent. 

\subsection{Training Objective and Optimization}
\label{sec: ASAM}
In this section, we describe the overall training objective and optimization strategy for TALAS.
Our design combines complementary objectives that promote semantic alignment and well-structured representations, together with an optimization scheme tailored to teacher-student distillation, with the goal of improving generalization under a capacity gap.

% While the mechanisms proposed in Sections~\ref{sec: TAMD} and \ref{sec: LASD} ensure structural alignment between the student and teacher, minimizing the distillation loss purely via Stochastic Gradient Descent (SGD) carries a risk. Cosine-based embedding distillation enforces a directional constraint for each training instance on the student representation manifold. When the intrinsic dimensionality of the student manifold is significantly lower than that of the teacher, these constraints become geometrically incompatible, forming an over-determined fitting problem at the manifold level. Prior work by \citet{stanton2021does} suggests that student models can achieve high fidelity (matching the teacher's outputs) while suffering from degraded generalization, a phenomenon attributed to the student overfitting to the teacher's specific instance rather than capturing the underlying data manifold. To mitigate this and bridge the capacity gap robustly, we integrate \textbf{Adaptive Sharpness-Aware Minimization (ASAM)}~\cite{kwon2021asamadaptivesharpnessawareminimization} into our training objective.

\paragraph{Overall Training Objective.}
The objectives introduced in Sections~\ref{sec: TAMD} and~\ref{sec: LASD} focus on aligning the student with the teacher’s semantic space.
However, exclusive reliance on such direct supervision carries a risk: if the teacher's representations exhibit collapse or high anisotropy within the specific training domain, the student may inherit these suboptimal geometric properties, potentially leading to a restricted embedding space \cite{ethayarajh-2019-contextual,li-etal-2020-sentence}.
In contrast, contrastive learning mitigates such collapse by optimizing for \textit{uniformity}~\citep{wang2020understanding}, which encourages distinct samples to separate and populate the hypersphere evenly.
This prevents the representation space from degenerating into anisotropic clusters, maximizing the model's discriminative power.
Therefore, we incorporate an unsupervised contrastive objective $\mathcal{L}_{\text{SimCSE}}$ based on SimCSE~\citep{gao2022simcsesimplecontrastivelearning} into our framework.
This term serves as a regularizer that complements the distillation process: it ensures the student benefits from the teacher's semantic guidance while simultaneously maintaining a broader, more uniform representation space, thereby enhancing generalization.

Given a batch of input sentences, we feed them through the student encoder twice with different standard dropout masks $z, z'$ to obtain two views of embeddings $\mathbf{e}_i^{z}$ and $\mathbf{e}_i^{z'}$. The loss is formulated as:
\begin{equation}
    \mathcal{L}_{\text{SimCSE}} = - \frac{1}{N} \sum_{i=1}^N \log \frac{e^{\text{sim}(\mathbf{e}_i^{z}, \mathbf{e}_i^{z'}) / \tau}}{\sum_{j=1}^N e^{\text{sim}(\mathbf{e}_i^{z}, \mathbf{e}_j^{z'}) / \tau}},
\end{equation}
where $N$ is the batch size, $\tau$ is the temperature hyperparameter, and $\text{sim}(\cdot)$ denotes cosine similarity.

The final training objective is defined as a weighted sum of three terms:
\begin{equation}
    \mathcal{L}_{\text{total}} = \lambda_1 \mathcal{L}_{\text{SimCSE}} + \lambda_2 \mathcal{L}_{\text{TAMD}} + \lambda_3 \mathcal{L}_{\text{LASD}},
    \label{eq:total_loss}
\end{equation}
where $\lambda_1, \lambda_2, \lambda_3$ are hyperparameters balancing the contributions of contrastive learning, teacher-anchored distillation, and self-distillation, respectively.
\paragraph{Optimization via ASAM.}
While the proposed objectives effectively align student representations with the teacher, optimizing the distillation loss using standard SGD can be risky under a teacher-student capacity gap.
Cosine-based embedding distillation imposes strong instance-level constraints, which may cause the student to overfit the teacher’s outputs without capturing the underlying data structure.
As shown by \citet{stanton2021does}, such overfitting can result in high fidelity to the teacher but poor generalization.

To mitigate this issue, we optimize $\mathcal{L}_{\text{total}}$ using Adaptive Sharpness-Aware Minimization (ASAM)~\citep{kwon2021asamadaptivesharpnessawareminimization}.
Following the standard ASAM formulation, we seek solutions that remain stable within a local neighborhood by solving the corresponding min-max problem in Eq.~\ref{eq:asam_talas}, which is efficiently optimized via a two-step ascent--descent procedure (Algorithm~\ref{alg:asam_talas}).

In our distillation setting, ASAM encourages convergence to flatter solutions, helping the student learn more robust and transferable representations under a capacity gap.

\begin{algorithm}[t]
\caption{TALAS Training via ASAM}
\label{alg:asam_talas}
\SetAlgoNoLine
\DontPrintSemicolon
\SetInd{0.6em}{0.01em}

\KwIn{Training corpus $\mathcal{D}$, Teacher $f_\mathcal{T}$, Student $f_\mathcal{S}(\cdot;\mathbf w)$, learning rate $\eta$, ASAM radius $\rho$}

\BlankLine
Pre-compute and cache teacher embeddings: 
$\mathbf{D}_e^\mathcal{T} \gets \{f_\mathcal{T}(x)_{\text{anchored}}\}_{x \in \mathcal{D}}$\;
\BlankLine
\For{each training step}{
    Sample mini-batch $\mathcal{B} \subset \mathcal{D}$\;
    Retrieve cached targets $\mathbf{E}^\mathcal{T} \in \mathbf{D}_e^\mathcal{T}$ for $\mathcal{B}$\;
    
    \BlankLine
    \tcp{1. Ascent Step}
    Compute standard loss following Equation: $\mathcal{L} \gets \mathcal{L}_{\text{total}}(f_\mathcal{S}(\mathcal{B}; \mathbf w), \mathbf{E}^\mathcal{T})$\;
    
    Generate perturbed weights $\tilde{\mathbf{w}}$:
    $\tilde{\mathbf{w}} \gets \text{ASAM\_Perturb}(\mathbf{w}, \nabla_{\mathbf w}\mathcal{L}, \rho)$\;
    
    \BlankLine
    \tcp{2. Descent Step}
    Compute gradients at the position $\tilde{\mathbf{w}}$:
    $\mathbf g_{\text{ASAM}} \gets \nabla_{\tilde{\mathbf{w}}}\mathcal{L}_{\text{total}}(f_\mathcal{S}(\mathcal{B}; \tilde{\mathbf{w}}), \mathbf{E}^\mathcal{T})$\;
    
    Update original parameters:
    $\mathbf w \gets \mathbf w - \eta \cdot \mathbf g_{\text{ASAM}}$\;
}
\end{algorithm}

\section{Experiments}
\begin{table*}[ht]
\centering
\resizebox{\textwidth}{!}{%
\begin{tabular}{l|ccc|ccc|ccc|cc|c}
\toprule
& \multicolumn{3}{c|}{\textbf{Classification (F1)}} 
& \multicolumn{3}{c|}{\textbf{Pair Classification (AP)}}
& \multicolumn{3}{c|}{\textbf{STS (Spearman)}}
& \multicolumn{2}{c|}{\textbf{Domain Avg}} \\
\cmidrule(lr){2-10}\cmidrule(lr){11-12}
& \textbf{Banking77} & \textbf{Tweet} & \textbf{Emotion$^{\star}$} 
& \textbf{MRPC} & \textbf{SciTail} & \textbf{WiC$^{\star}$} 
& \textbf{SICK} & \textbf{STS12} & \textbf{STSB$^{\star}$} 
& \textbf{Avg-In} & \textbf{Avg-Out} & \textbf{Avg.} \\
\midrule
\multicolumn{13}{c}{\textbf{Qwen3-Embedding 0.6B $\rightarrow$ MiniLMv2 H384 22M}} \\
\midrule
Teacher          
& 93.21 & 71.33 & 66.53 & 83.37 & 90.00 & 66.85 & 80.64 & 77.11 & 84.60 & 72.66 & 82.61 & 79.29 \\

Student base & 68.29 & 41.07 & 69.90 & 78.39 & 64.56 & 59.58 & 47.60 & 21.95 & 22.08 & 40.91 & 58.45 & 52.60 \\

SimCSE-unsup            
& 86.67 & 69.56 & 54.00 & 82.29 & 73.76 & 65.71 & 66.04 & 59.11 & 62.08 & 60.60 & 72.91 & 68.80 \\
\midrule

CDM            
& 86.07 & 70.32 & 53.04 & 81.97 & 72.97 & 68.34 & 65.97 & 61.12 & 65.75 & 62.38 & 73.07 & 69.51 \\
% CDM         
% & 86.16 & 70.21 & 53.68 & 81.81 & 73.46 & 68.35 & 66.02 & 61.26 & 65.87 & 62.63 & 73.15 & 69.65 \\
DSKD         
& 85.66 & 69.86 & 52.25 & 82.17 & 73.52 & 68.58 & 66.54 & 62.52 & 66.70 & 62.51 & 73.38 & 69.76 \\

Jasper and Stella     
& 85.94 & 71.25 & 57.99 & 83.36 & 76.17 & 68.25 & 70.57 & 63.81 & 70.11 & 65.45 & 75.18 & 71.94 \\
DistillCSE     
& \textbf{88.03} & 69.87 & 54.24 & 83.20 & 77.66 & 67.88 & 70.09 & 68.19 & 71.53 & 64.55 & 76.17 & 72.30 \\
EMO    
& 87.03 & 71.67 & 59.63 & 83.45 & 78.17 & \textbf{68.89} & 70.81 & 65.78 & 71.42 & 66.65 & 76.15 & 72.98 \\

\midrule
\rowcolor{gray!20}
\textbf{TALAS} 
& 86.69 & \textbf{72.77} & \textbf{60.94} & \textbf{85.10} & \textbf{81.42} & 66.65 
& \textbf{72.51} & \textbf{70.77} & \textbf{76.22} 
& \textbf{67.94} & \textbf{78.21} & \textbf{74.79} \\

\midrule
\multicolumn{13}{c}{\textbf{BGE-M3 $\rightarrow$ MiniLMv2 H768 66M}} \\
\midrule
Teacher          
& 93.52 & 73.85 & 68.56 & 85.81 & 91.87 & 61.47 & 79.18 & 78.73 & 84.87 & 71.63 & 83.83 & 79.76 \\

Student base & 81.84 & 47.70 & 71.67 & 79.67 & 65.06 & 60.89 & 49.46 & 26.76 & 24.12 & 44.24 & 62.41 & 56.35 \\

SimCSE-unsup                 
& 87.89 & 71.45 & 55.65 & 83.87 & 76.32 & 67.96 & 66.93 & 59.66 & 63.97 & 62.53 & 74.35 & 70.41 \\
\midrule
% CDM         
% & 88.95 & 70.80 & 59.45 & 83.67 & 75.85 & 70.02 & 67.70 & 62.64 & 67.41 & 65.63 & 74.94 & 71.83 \\

CDM            
& 89.15 & 70.48 & 58.71 & 84.16 & 75.91 & \textbf{70.00} & 68.14 & 62.74 & 67.78 & 65.50 & 75.10 & 71.90 \\
DSKD         
& 89.51 & 70.53 & 57.27 & 84.39 & 75.36 & 69.96 & 68.81 & 66.78 & 70.05 & 65.76 & 75.90 & 72.52 \\
Jasper and Stella      
& 88.38 & 74.13 & 62.81 & 85.98 & 80.33 & 68.38 & 74.35 & 66.80 & 74.55 & 68.58 & 78.33 & 75.08 \\
DistillCSE     
& 90.50 & 72.70 & 59.88 & 84.39 & 78.48 & 69.49 & 73.97 & 70.00 & 73.73 & 67.70 & 78.34 & 74.79 \\
EMO    
& 90.19 & 74.26 & 61.48 & 85.31 & 80.82 & 68.30 & 75.41 & 67.06 & 76.35 & 68.71 & 78.84 & 75.46 \\

\midrule
\rowcolor{gray!20}
\textbf{TALAS} 
& \textbf{90.62} & \textbf{74.81} & \textbf{63.30} & \textbf{87.05} & \textbf{81.94} & 66.28 & \textbf{76.86} & \textbf{69.04} & \textbf{79.01} & \textbf{69.53} & \textbf{80.05} & \textbf{76.55} \\

\midrule
\multicolumn{13}{c}{\textbf{Qwen3-Embedding 4B $\rightarrow$ Bert-base 109M}} \\
\midrule
Teacher          
& 93.63 & 72.22 & 68.94 & 84.79 & 91.77 & 66.80 & 83.43 & 82.55 & 86.87 & 74.20 & 84.73 & 81.22 \\

Student base & 84.86 & 47.79 & 68.02 & 77.36 & 67.74 & 59.22 & 42.43 & 21.54 & 20.30 & 42.44 & 60.33 & 54.36 \\

SimCSE-unsup           
& 89.21 & 69.13 & 53.55 & 82.86 & 76.59 & 71.64 & 68.19 & 61.68 & 70.36 & 65.18 & 74.61 & 71.47 \\
\midrule

CDM            
& 89.75 & 70.58 & 53.11 & 84.83 & 75.19 & 71.06 & 69.37 & 68.48 & 73.92 & 66.03 & 76.37 & 72.92 \\
% CDM         
% & 90.09 & 69.60 & 53.29 & 83.23 & 77.19 & 71.71 & 69.18 & 63.56 & 71.68 & 65.56 & 75.48 & 72.17 \\
DSKD         
& 89.66 & 69.93 & 54.51 & 83.13 & 77.95 & \textbf{71.53} & 68.65 & 63.37 & 71.42 & 65.82 & 75.45 & 72.24 \\

Jasper and Stella      
& 89.71 & 73.83 & 65.98 & 85.33 & 80.99 & 70.50 & 73.96 & 67.90 & 75.65 & 70.71 & 78.62 & 75.98 \\
DistillCSE     
& 90.94 & 70.90 & 60.80 & 82.86 & 78.98 & 68.63 & 75.12 & 72.61 & 77.08 & 68.84 & 78.57 & 75.32 \\
EMO    
& 90.78 & 73.54 & 67.48 & 84.46 & 81.57 & 69.76 & 75.66 & 68.83 & 77.17 & 71.47 & 79.14 & 76.58 \\

\midrule
\rowcolor{gray!20}
\textbf{TALAS} 
& \textbf{91.43} & \textbf{74.30} &\textbf{ 70.71} & \textbf{86.47} & \textbf{82.66} & 69.24 & \textbf{78.38} & \textbf{75.40} & \textbf{80.88} & \textbf{73.61} & \textbf{81.44} & \textbf{78.83} \\

\bottomrule
\end{tabular}
}
\caption{Experimental results on nine datasets. Datasets marked with $^{\star}$ are in-domain, while the remaining datasets are out-of-domain.}

\label{tab:main_results}
\end{table*}

\subsection{Experimental Setup}

\paragraph{Datasets.}
Our experiments span three task categories: \textbf{Classification}, \textbf{Pair Classification}, and \textbf{Semantic Textual Similarity (STS)}.
For \textbf{Classification}, we report F1 scores on \textsc{Banking77} \citep{casanueva2020banking77}, \textsc{Emotion} \citep{saravia2018emotion}, and \textsc{TweetEval-Sentiment} \citep{,barbieri2020tweeteval}.
For \textbf{Pair Classification}, we evaluate Average Precision (AP) on \textsc{MRPC} \cite{dolan2004mrpc}, \textsc{SciTail} \cite{khot2018scitail}, and \textsc{WiC} \citep{pilehvar2019wic}.
For \textbf{STS}, we report Spearman’s rank correlation on \textsc{SICK} \cite{marelli2014sick}, \textsc{STS12} \cite{agirre2012sts}, and \textsc{STS-Benchmark} \citep{cer2017stsb}.

We follow the task selection protocol of EMO \citep{truong2025emo} but differ in the training paradigm.
While EMO employs supervised fine-tuning (SFT), we adopt an \emph{unsupervised SimCSE-style training objective} \citep{gao2022simcsesimplecontrastivelearning} to encourage the student model to learn general-purpose semantic representations without relying on task-specific labels.

For training, we uniformly sample \textbf{5,000 sentences from each of three in-domain datasets} and merge them into a single unlabeled corpus that is shared across all teacher--student configurations.
Specifically, \textsc{Emotion}, \textsc{WiC}, and \textsc{STS-Benchmark} are treated as \emph{in-domain}, while the remaining datasets are considered \emph{out-of-domain}.
The resulting corpus is evaluated on all nine downstream benchmarks, yielding a balanced and domain-diverse training setup while strictly avoiding any form of supervised signal.

\paragraph{Training and Evaluation Settings.}
The student models considered in our experiments include \textsc{BERT}-base \citep{devlin2019bertpretrainingdeepbidirectional}, \textsc{MiniLMv2 H384}, and \textsc{MiniLMv2 H768} \citep{gu2024minillmknowledgedistillationlarge}.
For teacher models, we use \textsc{bge-m3} \citep{chen2024bgem3embeddingmultilingualmultifunctionality}, \textsc{qwen3-embedding-0.6B}, and \textsc{qwen3-embedding-4B} \citep{zhang2025qwen3embeddingadvancingtext}, and systematically evaluate multiple teacher~$\rightarrow$~student distillation configurations.

Detailed descriptions of the model architectures, data preprocessing procedures, training hyperparameters, and baseline implementations are provided in Appendix~\ref{appendix:dataset} and Appendix ~\ref{appendix:exp}.

\paragraph{Baselines.}
To rigorously evaluate the proposed framework, we compare it against representative knowledge distillation methods spanning different levels of granularity. 
For \textbf{token-level KD}, we include \textsc{DSKD}~\citep{zhang2024dualspaceknowledgedistillationlarge}, \textsc{CDM}~\citep{chen2025enhancingcrosstokenizerknowledgedistillation}, and \textsc{EMO}~\citep{truong2025emo}; these approaches transfer knowledge by aligning fine-grained token-level hidden representations or relational structures between teacher and student models. 
For \textbf{sentence-level KD}, we benchmark against \textsc{Jasper \& Stella}~\citep{zhang2025jasperstelladistillationsota} and \textsc{DistillCSE}~\citep{xu2023distillcse}, state-of-the-art methods that distill knowledge exclusively via global sentence embeddings without relying on token-wise supervision.

\subsection{Main Results}
To assess the effectiveness of \textit{TALAS}, we report the main experimental results with a primary focus on \textbf{embedding generalization}. Specifically, we evaluate whether the proposed method improves the robustness and transferability of sentence representations across diverse downstream tasks, thereby validating the semantic quality of the distilled student models compared to strong baselines. Comprehensive ablation studies and additional experimental analyses are provided in the Appendix \ref{appendix:exp}

\textbf{Table}~\ref{tab:main_results} shows that \textit{TALAS} consistently achieves the best or second-best performance across most datasets and attains the highest \textbf{Domain Average} score, indicating superior generalization on both in-domain and out-of-domain benchmarks. In particular, \textit{TALAS} yields notable improvements on challenging out-of-domain datasets such as \textit{Tweet} and \textit{STS12}, reflecting stronger robustness to domain shift. Compared with token-level distillation methods and sentence-level baselines, our approach delivers consistent gains without sacrificing efficiency, highlighting its ability to better preserve the geometric structure of the teacher embedding space while maintaining compact student representations.

% \textbf{Table}~\ref{tab:runtime_qwen4b_bert} reports the runtime and GPU memory footprint of different distillation methods under the \textbf{Qwen3-4B $\rightarrow$ BERT-base} setting. We measure the end-to-end training time per step, along with the peak, mean, and standard deviation of GPU memory consumption. Token-level distillation methods such as EMO, DSKD, and CDM incur substantial memory overhead, with peak usage exceeding 10~GB, primarily due to the heavy cost of teacher-side inference and token-wise alignment.

% In contrast, our method exhibits significantly improved computational efficiency. Specifically, \textit{TALAS (w/o ASAM)} reduces peak memory consumption from over 10~GB to 2.86~GB, corresponding to a reduction of more than \textbf{70}, while also achieving the fastest runtime among all compared approaches. When ASAM is enabled, \textit{TALAS (with ASAM)} introduces a moderate increase in computation and memory usage, yet still maintains a substantially lower footprint than existing baselines. These results demonstrate that our method achieves a favorable balance between representation quality and system efficiency, making it well suited for large-scale distillation under limited computational resources.

\section{Analysis}
\label{sec:analysis}

\begin{table}[!ht]
\centering
\footnotesize
\setlength{\tabcolsep}{1.5pt}
\renewcommand{\arraystretch}{1.1}
\begin{tabular}{c c c c | c c c}
\toprule
$\mathcal{L}_{\text{SimCSE}}$ & $\mathcal{L}_\text{TAMD}$ & $\mathcal{L}_{\text{LASD}}$ & ASAM 
& \textbf{Avg-In} & \textbf{Avg-Out} & \textbf{Avg-All} \\
\midrule
\multicolumn{7}{c}{\textit{Qwen3-Embedding 0.6B $\rightarrow$ MiniLMv2 H384 22M}} \\
\midrule
\checkmark &           &           &           & 44.24 & 62.41 & 56.35 \\
\checkmark & \checkmark&           &           & 65.50 & 75.24 & 71.99 \\
\checkmark & \checkmark& \checkmark&           & 67.03 & 75.77 & 72.85 \\
           & \checkmark&           & \checkmark& 67.65 & 77.10 & 73.95 \\
           & \checkmark& \checkmark& \checkmark& 67.27 & 77.78 & 74.28 \\
\checkmark & \checkmark&           & \checkmark& 68.04 & 76.93 & 73.97 \\
\checkmark & \checkmark& \checkmark& \checkmark& 67.94 & 78.21 & 74.79 \\

\midrule
\multicolumn{7}{c}{\textit{BGE-M3 0.6B $\rightarrow$ MiniLMv2 H768 66M}} \\
\midrule
\checkmark &           &           &           & 40.91 & 58.45 & 52.60 \\
\checkmark & \checkmark&           &           & 70.08 & 79.21 & 76.17 \\
\checkmark & \checkmark& \checkmark&           & 69.37 & 79.63 & 76.21 \\
           & \checkmark&           & \checkmark& 69.46 & 79.12 & 75.90 \\
           & \checkmark& \checkmark& \checkmark& 68.67 & 79.06 & 75.59 \\
\checkmark & \checkmark&           & \checkmark& 69.83 & 79.27 & 76.13 \\
\checkmark & \checkmark& \checkmark& \checkmark& 69.53 & 80.09 & 76.57 \\

\midrule
\multicolumn{7}{c}{\textit{Qwen3-Embedding 4B $\rightarrow$ Bert-base 109M}} \\
\midrule
\checkmark &           &           &           & 65.18 & 74.61 & 71.47 \\
\checkmark & \checkmark&           &           & 69.58 & 77.53 & 74.88 \\
\checkmark & \checkmark& \checkmark&           & 71.28 & 77.99 & 75.75 \\
           & \checkmark&           & \checkmark& 73.05 & 80.29 & 77.88 \\
           & \checkmark& \checkmark& \checkmark& 72.32 & 80.97 & 78.09 \\
\checkmark & \checkmark&           & \checkmark& 73.32 & 80.44 & 78.06 \\
\checkmark & \checkmark& \checkmark& \checkmark& 73.61 & 81.44 & 78.83 \\

\bottomrule

\end{tabular}

\caption{Ablation study on different loss combinations. Results are reported in terms of in-domain, out-of-domain, and overall average performance.}
\label{tab:loss_ablation_if}
\end{table}

\paragraph{Impact of each component in TALAS.}
$\mathcal{L}_\text{TAMD}$ functions as the primary bridge for cross-tokenizer alignment. As evidenced in Table~\ref{tab:loss_ablation_if}, configurations lacking $\mathcal{L}_\text{TAMD}$ suffer from a catastrophic performance drop, verifying its status as the fundamental backbone upon which other objectives operate. The integration of ASAM further elevates performance, yet its impact varies across configurations. While ASAM yields marginal gains for the \textit{BGE-M3 $\to$ MiniLMv2 H768} pair, it delivers substantial improvements for the \textit{Qwen3-Embedding 4B $\to$ BERT-base} and \textit{Qwen3-Embedding 0.6B $\to$ MiniLMv2 H384} settings. This disparity highlights ASAM's critical role in bridging the capacity gap: when the student is extremely compact (22M) or the teacher is vastly superior (4B), the optimization landscape becomes perilous, and ASAM's ability to seek flat minima proves essential for effective knowledge transfer.
Finally, the auxiliary objectives exhibit complementary roles: $\mathcal{L}_{\text{SimCSE}}$ enhances local clustering for in-domain tasks (Avg-In) but shows signs of bias toward the training distribution. Conversely, replacing it with $\mathcal{L}_{\text{LASD}}$ consistently improves out-of-domain generalization (Avg-Out), suggesting that preserving the geometric relationships between layers is more effective for robust transfer learning.

\paragraph{Computational Efficiency.}
Table~\ref{tab:runtime_qwen4b_bert} analyzes the resource consumption for the Qwen3-Embedding 4B $\rightarrow$ BERT-base distillation setting, measured on an NVIDIA T4 GPU with a batch size of 32. Token-level baselines (e.g., EMO, DSKD) incur prohibitive overheads with peak memory exceeding 12~GB due to the necessity of parallel teacher inference and the impracticality of caching dense states. Among lightweight alternatives, \textit{Jasper} shows low latency but requires a cumbersome two-stage pipeline, while \textit{DistillCSE}, although faster, still consumes more memory and lacks structural guidance from intermediate representations. In contrast, our framework achieves the optimal synergy of efficiency and performance. TALAS (w/o ASAM) records the \emph{second fastest} distillation speed of 195 ms/step—slightly behind DistillCSE (182 ms/step) and outperforming Jasper (204 ms/step)—while still delivering highly competitive results with a minimal memory footprint (3.98 GB). Even with ASAM enabled, TALAS remains significantly more resource-friendly than token-level methods, making it the most viable solution for high-performance distillation on consumer-grade hardware.
\begin{table}[h]
\centering
\scriptsize
\resizebox{0.49\textwidth}{!}{
\begin{tabular}{l c c c c}
\toprule
Method & Time & Mean Mem & Peak Mem & Std Mem \\
       & (ms/step) & (MB) & (MB) & (MB) \\
\midrule
SimCSE-unsup                & 151  & 2662 & 3690 & 235 \\
DistillCSE                & 182  & 3373 & 3674 & 124 \\
Jasper and Stella   & 204 & 2932 & 3884 & 71 \\
% Sentence Embedding& 186  & 2873  & 3894  & 265 \\
EMO     & 1223 & 11203 & 13831 & 492 \\
DSKD    & 1002 & 10991 & 12393 & 336 \\
CDM     & 1078 & 10814 & 12156 & 314 \\
TALAS (w/o ASAM)        & 195 & 2864 & 3982 & 261 \\
TALAS (with ASAM)       & 375  & 3317 & 4499 & 277 \\
\bottomrule
\end{tabular}}
\caption{Runtime and GPU memory while training comparison for Qwen3-Embedding-4B $\rightarrow$ BERT-base.}
\label{tab:runtime_qwen4b_bert}
\end{table}

\begin{figure}[h]
\centering
\includegraphics[width=\linewidth]{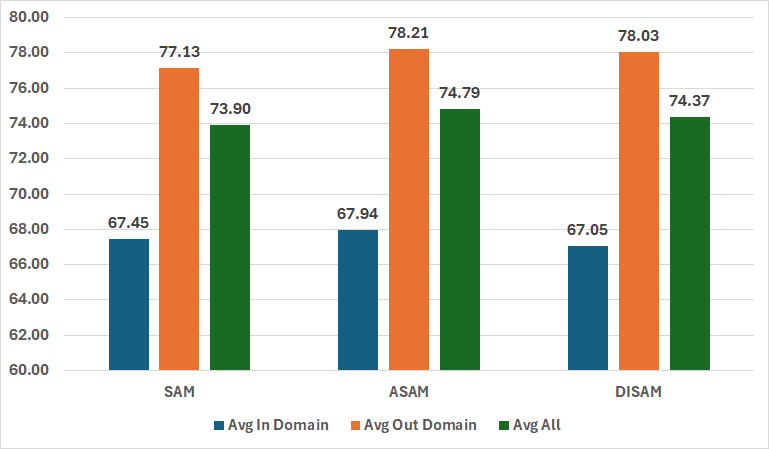}
\captionof{figure}{Comparison of SAM variants (SAM, DISAM, and ASAM) for Qwen3-Embedding 0.6B $\rightarrow$ MiniLMv2 H384.}
\label{fig:SAM_abl}
\end{figure}

\paragraph{Impact of Sharpness-Aware Minimization Variants.}
Figure~\ref{fig:SAM_abl} presents the comparative results of SAM, ASAM, and DISAM on the Qwen3-Embedding 0.6B $\to$ MiniLMv2 H384 distillation task (refer to Appendix \ref{appendix:addition_ablation} for detailed breakdowns on individual datasets). As illustrated, ASAM consistently achieves the superior performance, recording the highest overall average score of \textbf{74.79} compared to DISAM (74.37) and SAM (73.90). Specifically, ASAM demonstrates remarkable robustness in Out-of-Domain generalization (78.21) while maintaining the best In-Domain stability (67.94). Consequently, we identify ASAM as the optimal optimizer for TALAS to ensure a robust balance between generalization capabilities and representation fidelity.

\paragraph{Empirical Sharpness Analysis.}
To further substantiate the role of sharpness in our setting, we directly measure curvature during training by estimating the largest eigenvalue of the Hessian, denoted as $\lambda_{\max}$. This quantity serves as a widely adopted proxy for loss landscape sharpness, where larger values indicate sharper minima and smaller values correspond to flatter regions.

We track $\lambda_{\max}$ of the full training objective across epochs for both ASAM and AdamW under identical training conditions. The results are summarized in Table~\ref{tab:hessian_efficiency}.

\begin{table}[h]
\centering
\begin{tabular}{c|cc}
\hline
\textbf{Epoch} & \textbf{ASAM} & \textbf{AdamW} \\
\hline
1  & 14.89 & 30.24 \\
3  & 12.94 & 21.48 \\
5  & 11.59 & 19.66 \\
10 & 10.85 & 18.66 \\
30 & 9.93  & 17.41 \\
\hline
\end{tabular}
\caption{Largest Hessian eigenvalue ($\lambda_{\max}$) across training epochs. Lower values indicate flatter minima.}
\label{tab:hessian_efficiency}
\end{table}

As shown in Table~\ref{tab:hessian_efficiency}, ASAM consistently produces significantly smaller $\lambda_{\max}$ values compared to AdamW at all measured epochs. Notably, the gap remains substantial throughout training, indicating that ASAM systematically steers optimization toward flatter regions of the loss landscape rather than merely converging to a similar solution with reduced variance.

This behavior aligns with the theoretical motivation of sharpness-aware methods, which explicitly penalize sensitivity to parameter perturbations. By minimizing the worst-case loss within a local neighborhood, ASAM effectively avoids sharp minima characterized by high curvature. In contrast, AdamW, which lacks such regularization, tends to converge to sharper regions with higher $\lambda_{\max}$.

\begin{figure}[h]
\centering
\includegraphics[width=\linewidth]{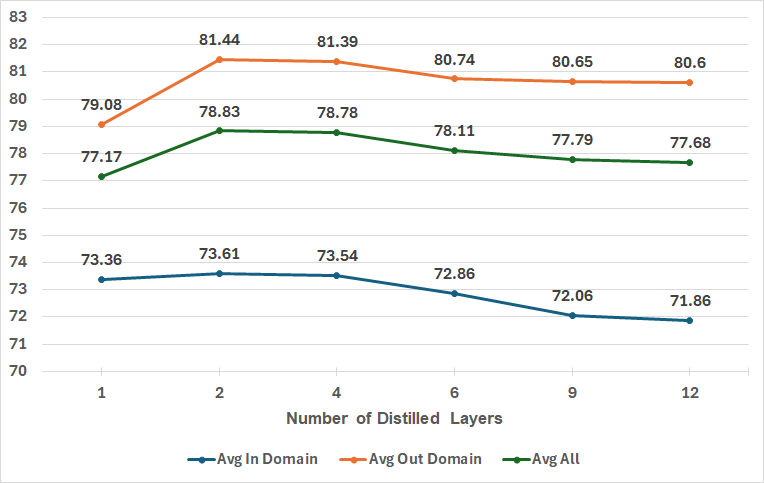}
\captionof{figure}{Effect of the number of distilled layers from teacher for Qwen3-Embedding 4B $\rightarrow$ BERT.}
\label{fig:num_distil}
\end{figure}

\paragraph{Effect of Distilled Layer Depth.}
Figure \ref{fig:num_distil} illustrates the impact of the number of teacher-anchored layers ($K$) on performance. We observe that effectiveness peaks at $K=2$ (Avg All \textbf{78.83}) rather than scaling linearly, with consistent degradation occurring at deeper alignments ($K > 4$). This trend suggests that given the substantial teacher-student capacity gap, enforcing constraints on too many internal layers induces over-regularization, restricting the student's intrinsic feature organization. Consequently, anchoring only the top two layers yields the optimal balance between high-level semantic transfer and the student's representational flexibility.

% \paragraph{Comparison of SAM variants}

% \paragraph{Comparison of Domain Average}

\section{Conclusion}
We presented \textbf{TALAS}, a novel distillation framework that effectively compresses LLM capabilities into compact encoders via a dual-strategy approach: aligning upper layers with the teacher ($\mathcal{L}_{\text{TAMD}}$) and enforcing internal geometric consistency ($\mathcal{L}_{\text{LASD}}$). Furthermore, we demonstrated that integrating ASAM is critical for navigating the student's loss landscape, enabling convergence to flat minima that generalize well across domains. Experimental results verify that TALAS outperforms existing baselines in both in-domain and out-of-domain scenarios. Future work will focus on scaling the training data and extending our evaluation to the MTEB to assess the model's versatility.

\section{Limitations}
\label{sec:limitations}

Despite the effectiveness of TALAS, we acknowledge certain limitations in our study. First, a constrained computational budget restricted our experimental scope, precluding a comprehensive investigation into the framework's scalability with large models or large-scale training corpora. Second, our current evaluation focuses primarily on English-centric benchmarks; thus, the framework's adaptability to multilingual settings or low-resource languages remains to be fully explored. Finally, while our method significantly reduces memory footprint, the integration of ASAM necessitates two forward-backward passes per update step, effectively doubling the training duration. This trade-off between convergence stability and training speed may present a constraint for scenarios requiring rapid iteration cycles or when scaling to very large student models where temporal efficiency is paramount.

\section*{Acknowledgments}
Trung Le was supported by the Air Force Office of Scientific Research under award number FA2386-25-1-4023 and the ARC Discovery Project grant DP250100262. Thien Huu Nguyen was supported by NSF Grant \#2239570. He is also supported in part by the Office of the Director of National Intelligence (ODNI), Intelligence Advanced Research Projects Activity (IARPA), via the HIATUS Program contract 2022-22072200003. The views and conclusions contained herein are those of the authors and should not be interpreted as necessarily representing the official policies, either expressed or implied, of ODNI, IARPA, or the U.S. Government.

\bibliography{ref}
\bibliographystyle{acl_natbib}

\newpage
\appendix
\newpage
% \onecolumn
\appendix

\section{Extended Related Work}
\label{app:erw}

Knowledge Distillation (KD) aims to transfer knowledge from a high-capacity teacher model to a compact student model, enabling efficient deployment while preserving performance.
Since its introduction, KD has been extensively studied across a wide range of NLP tasks, including classification, text generation, and representation learning.
Despite its success, effective distillation remains challenging when compressing large-scale models into small students, particularly under limited computational and memory budgets.

\paragraph{Output Distillation}
Output distillation is the most widely adopted form of knowledge distillation, where the student is trained to match the teacher’s output distributions or representations.
Early work focuses on distilling softened logits or output embeddings, encouraging the student to imitate the teacher’s predictive behavior at the output level \citep{hinton2015distillingknowledgeneuralnetwork,sanh2020distilbertdistilledversionbert,jiao2020tinybertdistillingbertnatural, vu2026dwa}. In representation learning, output-level distillation is especially attractive due to its simplicity and efficiency: it does not require access to the teacher’s internal states and naturally supports offline or black-box distillation scenarios \citep{palo-etal-2024-performance,agarwal2024onpolicydistillationlanguagemodels,gu2024minillmknowledgedistillationlarge}.
However, relying solely on output supervision can be insufficient when the capacity gap between the teacher and student is large, often leading to suboptimal generalization or overfitting to teacher-specific artifacts \citep{cho2019efficacyknowledgedistillation,ko2024distillmstreamlineddistillationlarge}.
These limitations motivate more structured forms of knowledge transfer beyond point-wise imitation.

\paragraph{Relational Distillation}
To address the shortcomings of output-level supervision, relational distillation methods aim to transfer structural information among samples rather than matching individual representations.
Relational Knowledge Distillation (RKD) preserves pairwise distances or angular relationships in the embedding space, demonstrating improved robustness and generalization, particularly in metric learning settings \citep{park2019relationalknowledgedistillation}. In NLP, relational signals have been leveraged to compress large Transformer models by distilling attention patterns or hidden-state correlations, showing that preserving semantic geometry can be more effective than strict feature-level alignment \citep{jiao2020tinybertdistillingbertnatural,sun2020mobilebertcompacttaskagnosticbert,huang2023fcd,truong2025emo, hoang2026mcw}.
While effective, these approaches often incur substantial memory and computational overhead due to pairwise comparisons or intermediate-state supervision, which limits scalability - especially when teacher inference must be performed online during training.

\paragraph{Teacher Assistant Distillation}
Another line of work addresses the difficulty of distilling across large capacity gaps by introducing intermediate models between the teacher and the final student.
Teacher Assistant (TA) distillation progressively transfers knowledge through one or more assistant models with intermediate capacities, enabling smoother knowledge transfer \citep{mirzadeh2020improved}.
This strategy has been shown to consistently outperform direct teacher-to-student distillation, particularly when the teacher is significantly larger than the student \citep{mirzadeh2020improved,son2021gkd}. Recent studies further explore staged or curriculum-based distillation strategies in large-scale or embedding-focused scenarios \citep{lin2023prodprogressivedistillationdense,han2024amdautomaticmultistepdistillation}.
However, TA-based methods substantially increase training complexity and computational cost due to the need to train and maintain additional intermediate models, which limits their practicality in resource-constrained environments.

\paragraph{Knowledge Distillation for Embedding Models}
Text embedding models serve as a foundational component across a wide range of NLP tasks, including semantic retrieval and retrieval-augmented generation \citep{hieu2025magix, nguyen2025improving}, event detection \citep{hai2026mozila, le2024sharpseq}, and relation extraction \citep{dao2026wave++, anh2025mutual, tran2024preserving, le2025enhancing, pham2025mitigating}. However, while knowledge distillation has recently been widely adopted to transfer knowledge across diverse settings - from reasoning capability transfer in large language models \citep{gu2024minillmknowledgedistillationlarge, ko2024distillmstreamlineddistillationlarge} to preference alignment \citep{truong2026ctpd, le2025token} - its application to text embedding models, despite their central role in these downstream tasks, remains substantially underexplored.

However, while recently knowledge distillation has been widely adopted to transfer knowledge across a range of tasks - from reasoning capability transfer in large language models \citep{gu2024minillmknowledgedistillationlarge, ko2024distillmstreamlineddistillationlarge} to preference alignment \cite{truong2026ctpd, le2025token} - yet its application to text embedding models remains substantially underexplored.
Distilling embedding models poses challenges that differ substantially from those encountered in standard classification or text generation tasks. High-performing embedding teachers are often large language models with architectures, tokenizers, or vocabularies that differ from those of the student, making token-level alignment either computationally expensive or infeasible. As a result, relatively few studies explicitly focus on embedding-specific distillation. Representative early works include \citet{Zhuang2021EnsembleDistillation}, which directly mimics output logits over the vocabulary, and \citet{jiao2020tinybertdistillingbertnatural}, which distills knowledge from both hidden representations and attention matrices. DistillCSE~\citep{xu2023distillcse} takes an important step toward embedding-level distillation by transferring sentence representations through contrastive learning, thereby avoiding token-level supervision and enabling more efficient knowledge transfer. More recently, Jasper and Stella~\citep{zhang2025jasperstelladistillationsota} propose a multi-stage distillation pipeline to compress state-of-the-art embedding models while maintaining strong retrieval performance. EMO~\citep{truong2025emo} further explores embedding distillation from a relational perspective, aligning inter-sample structures via optimal transport. More recently, \citet{an2026mol} proposes a Mixture-of-Layers approach for cross-tokenizer embedding distillation, adaptively routing distillation signals across multiple student layers to bridge the representational gap between heterogeneous teacher–student pairs.

Despite their effectiveness, many existing methods still rely on token-level supervision, intermediate-state matching, or repeated online teacher inference, which introduces significant memory and computational overhead during student training.

In contrast, our work focuses on \emph{resource-efficient embedding distillation}.
We distill knowledge using only cached sentence-level embeddings from the teacher, eliminating the need for teacher inference during student training.
This design significantly reduces memory consumption and training latency while remaining effective in large-teacher–small-student settings, making it well suited for practical deployment scenarios.

\section{Sharpness-Aware Optimization}
\label{appendix:sam}

\paragraph{Sharpness-Aware Minimization (SAM)  \citep{foret2020sharpness}.}
Standard empirical risk minimization tends to converge to sharp minima, which are sensitive to parameter perturbations and often generalize poorly. SAM addresses this by simultaneously minimizing the loss value and the loss landscape geometry (sharpness). Formally, it seeks parameters $\boldsymbol{\theta}$ that minimize the worst-case loss within a local Euclidean neighborhood of radius $\rho$:
\begin{equation}
    \min_{\boldsymbol{\theta}} \max_{\|\boldsymbol{\epsilon}\|_2 \leq \rho} \mathcal{L}(\boldsymbol{\theta} + \boldsymbol{\epsilon}).
\end{equation}
To solve the inner maximization efficiently, SAM applies a first-order Taylor approximation, deriving the optimal perturbation $\hat{\boldsymbol{\epsilon}}$ as the gradient direction scaled by $\rho$:
\begin{equation}
    \hat{\boldsymbol{\epsilon}} = \rho \frac{\nabla_{\boldsymbol{\theta}} \mathcal{L}(\boldsymbol{\theta})}{\|\nabla_{\boldsymbol{\theta}} \mathcal{L}(\boldsymbol{\theta})\|_2}.
\end{equation}
The model then updates its weights based on the gradient computed at the perturbed position $\boldsymbol{\theta} + \hat{\boldsymbol{\epsilon}}$, effectively steering the optimization trajectory towards flat minima that improve generalization.

\paragraph{Domain-Inspired Sharpness-Aware Minimization (DISAM) \citep{zhang2024domain}.}
Standard SAM operates under the crucial assumption that training samples are independent and identically distributed (i.i.d.). However, in multi-domain scenarios, this assumption often fails as data varies significantly in complexity and distribution across domains. Consequently, SAM suffers from \textit{inconsistent convergence}, where it generates perturbations biased towards domains with larger gradients (typically those that are under-converged), potentially destabilizing the optimization for other domains.

To rectify this, DISAM integrates domain-level statistics into the sharpness estimation by imposing a variational constraint. The objective is to maximize the perturbed loss while simultaneously minimizing the divergence between domain losses. Formally, the maximization step in DISAM is defined as:
\begin{equation}
\begin{split}
\mathcal{L}_{\text{DISAM}}(\mathbf{w}) 
= \max_{\|\boldsymbol{\epsilon}\|_2 \leq \rho} \Bigg[
& \sum_{i=1}^M \alpha_i \mathcal{L}_i(\mathbf{w} + \boldsymbol{\epsilon}) \\
& {}- \lambda \,\mathrm{Var}\{\mathcal{L}_i(\mathbf{w} + \boldsymbol{\epsilon})\}_{i=1}^M
\Bigg],
\end{split}
\end{equation}
where $\lambda$ is a hyperparameter balancing the sharpness and the consistency constraints. Explicitly, the inter-domain variance term is defined as
\begin{equation}
V(\mathbf{w} + \boldsymbol{\epsilon})
= \mathrm{Var}\{\mathcal{L}_i(\mathbf{w} + \boldsymbol{\epsilon})\}_{i=1}^M.
\end{equation}

It is calculated as the mean squared difference between domain losses:
\begin{equation}
\begin{split}
V(\mathbf{w} + \boldsymbol{\epsilon})
= \frac{1}{2M^2} \sum_{i=1}^M \sum_{j=1}^M 
\Big(
& \mathcal{L}_i(\mathbf{w} + \boldsymbol{\epsilon}) \\
& {}- \mathcal{L}_j(\mathbf{w} + \boldsymbol{\epsilon})
\Big)^2.
\end{split}
\end{equation}
Intuitively, this variance term acts as an adaptive regularizer. If a specific domain's loss deviates significantly from the others (violating the balance), the variance gradient counteracts the standard SAM perturbation. This mechanism essentially suppresses noise for under-converged domains (to maintain stability) while amplifying it for well-converged domains (to escape sharp minima), thereby synchronizing convergence across diverse data distributions.
Our work integrates adaptive sharpness-aware optimization directly into the embedding distillation pipeline.
By optimizing the distillation objective under ASAM, the student is encouraged to capture the teacher’s broad semantic structure rather than overfitting to specific embedding targets.
This combination of embedding-level distillation and sharpness-aware optimization is particularly effective for large teacher–small student scenarios, leading to improved robustness and out-of-domain generalization.
\section{Offline Teacher Inference and Cached Distillation Targets}
\label{appendix:offline_teacher}

The teacher model is not invoked during student training. Instead, it is executed \textbf{only once} in a preprocessing stage to extract and cache sentence-level representations for all training samples.

Specifically, we pass the training data through the teacher model using a dedicated dataloader and store the resulting last CLS embeddings associated with each sample. This procedure ensures full consistency between the teacher and student inputs, while avoiding discrepancies caused by differences in preprocessing or data ordering.

During student training, the cached distillation targets are retrieved alongside the input data through a \textit{lazy loading} mechanism. For each mini-batch, only the corresponding precomputed sentence embeddings are loaded from memory or disk, eliminating the need to perform teacher inference within the training loop. This design completely removes the computational overhead of the teacher during training and substantially reduces both training latency and GPU memory consumption compared to conventional online distillation approaches.

\textbf{Rationale.}

Consider a BERT-based teacher model, where each input sample produces hidden states of shape $[256, 768]$. This corresponds to
\[
256 \times 768 = 196{,}608 \text{ elements per sample}.
\]
When stored in \texttt{float32} format, this requires approximately $786{,}432$ bytes ($\approx 768$ KB) per sample. For a dataset of $100{,}000$ samples, the total storage requirement exceeds $76$ GB, which is impractical for most training environments. Approaches that rely on caching intermediate-layer hidden states further amplify this cost, rendering offline storage largely infeasible in practice.

In contrast, by caching only the \textit{sentence embedding}, each sample requires a single vector of shape $[768]$. When stored in \texttt{float32} format, this amounts to $3{,}072$ bytes ($\approx 3$ KB) per sample. Consequently, for $100{,}000$ samples, the total storage footprint is reduced to approximately $300$ MB, making offline distillation targets both practical and scalable.

\section{Dataset Creation}
\label{appendix:dataset}
To construct the training data for contrastive sentence representation learning, we sample sentences from multiple task categories to ensure diversity across domains and objectives. Specifically, we select 5,000 sentences from classification datasets, 2,500 sentence pairs from semantic textual similarity (STS) datasets, and 2,500 sentence pairs from pair classification datasets. All sentence pairs are then flattened into individual sentences, resulting in a total of 15,000 unique sentences. This unified sentence collection is used to form the training data for SimCSE-style contrastive learning, enabling the model to learn robust sentence embeddings from heterogeneous sources while maintaining a consistent training format.
\section{Experimental Details}
\label{appendix:exp}

\paragraph{Training Configuration}
We adopt a unified training protocol across the three teacher–student pairs considered in our experiments. The detailed training configurations for all methods are summarized in Table~\ref{tab:training-config}. All knowledge distillation (KD) experiments are conducted on the same dataset constructed as described in Section~\ref{appendix:dataset}, ensuring a fair comparison across methods. Evaluation is performed on multiple benchmark datasets spanning different domains and task types, using standard metrics including F1 score for classification, Average Precision (AP) for pair classification, and Spearman’s rank correlation for semantic textual similarity.

\begin{table*}[htbp]
\centering
\small
\begin{tabular}{l|cccccc}
\toprule
\textbf{Settings} 
& \textbf{DSKD} 
& \textbf{CDM} 
& \textbf{EMO} 
& \textbf{Stella \& Jasper} 
& \textbf{DistillCSE} 
& \textbf{TALAS} \\
\midrule
Epoch & 5 & 5 & 5 & 2 + 5 (two-stage) & 5 & 5 \\
LR                 & $2\times10^{-5}$ & $2\times10^{-5}$ & $2\times10^{-5}$ & $2\times10^{-5}$ & $2\times10^{-5}$ & $2\times10^{-5}$ \\
Batch Size         & 32  & 32  & 32  & 32  & 32  & 32 \\
LR Scheduler       & Cosine & Cosine & Cosine & Cosine & Cosine & Cosine \\
\bottomrule
\end{tabular}
\caption{Training configurations for different distillation and sentence embedding methods.}
\label{tab:training-config}
\end{table*}

\paragraph{Losses}
Our models were trained with $\lambda_1 = 0.001, \lambda_2 = 0.75 \ and \ \lambda_3 = 1$

\paragraph{Evaluation} We evaluate the quality of the learned sentence embeddings across three categories of downstream tasks. 
(1) For \textbf{classification tasks}, we follow standard evaluation settings \citep{conneau2018sentevalevaluationtoolkituniversal} by freezing the sentence embeddings and training a Logistic Regression classifier on top. 
(2) For \textbf{pair classification}, we compute the cosine similarity between sentence representations and determine the prediction based on an optimal threshold, reporting the Average Precision (AP). 
(3) Finally, for \textbf{Semantic Textual Similarity (STS)} tasks, we measure the alignment between the cosine similarity of the embeddings and human-annotated gold scores using Spearman correlation.

\paragraph{Baseline Implementation Details}
To ensure a rigorous evaluation, we adapt state-of-the-art generative distillation methods to the discriminative sentence embedding setting.

\noindent\textbf{Contextual Dynamic Mapping (CDM).} 
We apply CDM to align hidden representations $\mathbf{H} \in \mathbb{R}^{B \times L \times D}$. The original formulation relies on the teacher's predictive entropy to modulate the Dynamic Time Warping (DTW) alignment cost. However, since standard sentence encoders do not readily output token-level vocabulary distributions, we modify the cost function to utilize \textit{token edit distance} (Levenshtein distance). Specifically, we compute a pairwise distance matrix between the student and teacher token sequences and employ DTW to identify the optimal monotonic alignment path that minimizes the cumulative cost. Based on this path, teacher hidden states are aggregated (via averaging) to match the student's sequence length, establishing a synchronized one-to-one correspondence for the final Mean Squared Error (MSE) loss.

\noindent\textbf{Dual Space Knowledge Distillation (DSKD).} 
We extend DSKD to enforce geometric consistency through a symmetric, bidirectional process. Unlike standard unidirectional approaches, DSKD projects representations into each model's respective latent space. We implement this at two granularities: 
(1) \textit{Sentence-level:} Global representations (e.g., \texttt{[CLS]} tokens) are distilled via learnable linear projection heads to align the global semantic spaces. 
(2) \textit{Token-level:} To address the misalignment caused by differing vocabularies and sequence lengths, we incorporate Cross-Layer Attention (CLA)~\citep{Mu2024CrossLayerAttentionSharing}. CLA computes an attention mechanism between the student's and teacher's intermediate layers, allowing the student to reconstruct fine-grained teacher features (and vice versa) without requiring rigid token-to-token matching.

\section{Additional Experiment Results}
\label{appendix:addition_ablation}

\paragraph{Comparison with Existing Pretrained-Model.}

\begin{table*}[htbp]
\centering
\resizebox{\textwidth}{!}{%
\begin{tabular}{l|c|cc|cc|cc|c}
\toprule
& 
& \multicolumn{2}{c|}{\textbf{Classification (F1)}} 
& \multicolumn{2}{c|}{\textbf{Pair Classification (AP)}}
& \multicolumn{2}{c|}{\textbf{STS (Spearman)}}
& \textbf{Domain Avg} \\
\cmidrule(lr){3-4}\cmidrule(lr){5-6}\cmidrule(lr){7-8}
\textbf{Model} 
& \textbf{Training Data}
& \textbf{Banking77} & \textbf{Tweet}
& \textbf{MRPC} & \textbf{SciTail}
& \textbf{SICK} & \textbf{STS12}
& \textbf{Avg-Out} \\
\midrule
% princeton-nlp/unsup-simcse-bert-base-uncased  
Unsup-SimCSE
& $\sim$1M sentences
& 90.74 & 69.85
& 84.85 & 83.04
& 72.23 & 68.40
& 78.19 \\
% princeton-nlp/sup-simcse-bert-base-uncased 
Sup-SimCSE
& $\sim$1M pairs (NLI)
& \textit{90.77} & 74.10
& 86.76 & \textbf{85.40}
& \textbf{79.52} & \textit{77.76}
& \textbf{82.39} \\
% sentence-transformers/bert-base-nli-mean-tokens
SBERT-NLI (mean)
& $\sim$1M pairs (NLI)
& 88.84 & 74.00
& 86.97 & 76.67
& 69.71 & 70.15
& 77.72 \\
% sentence-transformers/bert-base-nli-stsb-mean-tokens  
SBERT-NLI+STS (mean)
& $\sim$1M pairs (NLI) + 8.6k (STS-B)
& 89.84 & 74.11
& \textbf{89.28} & \textit{83.76}
& 77.15 & \textbf{79.38}
& \textit{82.25} \\
SBERT-NLI (CLS)
& $\sim$1M pairs (NLI)
& 89.11 & \textbf{75.90}
& \textit{87.80} & 77.63
& 72.21 & 69.41
& 78.68 \\
\rowcolor{gray!20}
\textbf{TALAS} 
& $\sim$15K sentences
& \textbf{91.43} & \textit{74.30}
& 86.47 & 82.66
& \textit{78.38} & 75.40
& 81.44 \\
\bottomrule
\end{tabular}
}
\caption{Out-of-domain performance comparison with existing fine-tuned Bert-base model. Best results are shown in \textbf{bold}, and second-best results are shown in \textit{italics}.}
\label{tab:out_domain_results}
\end{table*}

Table~\ref{tab:out_domain_results} highlights that our method achieves strong out-of-domain generalization despite being trained on a substantially smaller dataset. While most competing baselines rely on large-scale supervision with approximately 1M sentence pairs or sentences, our approach is trained on only~15K sentences yet attains an Avg-Out score of 81.44, remaining competitive with state-of-the-art supervised models such as Sup-SimCSE (82.39) \citep{gao2022simcsesimplecontrastivelearning} and SBERT NLI+STS (82.25) \citep{reimers2019sentence}.

Across individual tasks, our model demonstrates consistent robustness under domain shift, achieving competitive results in both classification and semantic similarity benchmarks. Notably, our method performs strongly on STS tasks, indicating that effective semantic alignment can be learned even under limited-data conditions. This suggests that the proposed training strategy leverages teacher supervision efficiently, enabling high-quality sentence representations without dependence on large annotated corpora.

Overall, the table highlight the strong data efficiency and practical applicability of our method. Despite being trained on substantially fewer samples, our approach achieves competitive out-of-domain performance while incurring lower training time and memory costs compared to existing knowledge distillation methods. This makes our method a compelling alternative to fine-tuned BERT-base models that rely on large-scale supervised datasets.

\begin{table*}[h]
\centering
\resizebox{\textwidth}{!}{%
\begin{tabular}{l|ccc|ccc|ccc|cc|c}
\toprule
& \multicolumn{3}{c|}{\textbf{Classification (F1)}} 
& \multicolumn{3}{c|}{\textbf{Pair Classification (AP)}}
& \multicolumn{3}{c|}{\textbf{STS (Spearman)}}
& \multicolumn{2}{c|}{\textbf{Domain Avg}} \\
\cmidrule(lr){2-10}\cmidrule(lr){11-12}
\textbf{Num Layers} & \textbf{Banking77} & \textbf{Tweet} & \textbf{Emotion$^{\star}$} 
& \textbf{MRPC} & \textbf{SciTail} & \textbf{WiC$^{\star}$} 
& \textbf{SICK} & \textbf{STS12} & \textbf{STSB$^{\star}$} 
& \textbf{Avg-In} & \textbf{Avg-Out} & \textbf{Avg.} \\
\midrule
\multicolumn{13}{c}{\textbf{Qwen3-Embedding 0.6B $\rightarrow$ MiniLMv2 H384}} \\
\midrule
1            
& 86.28 & 72.33 & 60.46 
& 84.33 & 80.73 & 66.57 
& 72.15 & 68.14 & 75.43 
& 67.49 & 77.33 & 74.05 \\
2            
& 86.69 & 72.77 & 60.94 
& 85.10 & 81.42 & 66.65 
& 72.51 & 70.77 & 76.22 
& 67.94 & 78.21 & 74.79 \\
3
& 86.03 & 72.35 & 61.01 
& 85.10 & 80.55 & 66.86 
& 73.01 & 68.60 & 75.76 
& 67.88 & 77.61 & 74.36 \\
6     
& 85.95 & 71.72 & 60.88 
& 85.33 & 79.47 & 67.12 
& 71.85 & 67.33 & 74.13 
& 67.38 & 76.94 & 73.75 \\
\midrule
\multicolumn{13}{c}{\textbf{Qwen3-Embedding 4B $\rightarrow$ Bert-base}} \\
\midrule
1            
& 91.10 & 72.88 & 70.73 
& 85.36 & 82.30 & 71.27 
& 77.40 & 65.41 & 78.09 
& 73.36 & 79.08 & 77.17 \\
2            
& 91.43 & 74.30 & 70.71 
& 86.47 & 82.66 & 69.24 
& 78.38 & 75.40 & 80.88 
& 73.61 & 81.44 & 78.83 \\
4
& 91.35 & 73.99 & 70.94 
& 86.55 & 82.80 & 69.11 
& 77.80 & 75.86 & 80.58 
& 73.54 & 81.39 & 78.78 \\
6     
& 91.24 & 74.09 & 69.59 
& 86.45 & 82.48 & 69.26 
& 77.25 & 72.91 & 79.72 
& 72.86 & 80.74 & 78.11 \\
9        
& 90.05 & 74.84 & 69.10 
& 86.73 & 82.35 & 67.99 
& 76.76 & 73.17 & 79.09 
& 72.06 & 80.65 & 77.79 \\
12        
& 90.28 & 74.68 & 68.47 
& 86.74 & 82.99 & 68.15 
& 76.45 & 72.43 & 78.96 
& 71.86 & 80.60 & 77.68 \\

\bottomrule
\end{tabular}
}
\caption{Ablation on $\mathcal{L}_\text{TAMD}$ }

\label{tab:ablation_tamd}
\end{table*}

\paragraph{Ablation on Teacher Distillation.}
In table~\ref{tab:ablation_tamd}, for Qwen3-Embedding 0.6B → MiniLMv2 H384, increasing the number of distilled layers from 1 to 2 improves both Avg-Out and overall performance, with the peak Avg-Out achieved at 2 layers (78.21). Further increasing the number of distilled layers leads to gradual performance degradation, particularly on STS and out-of-domain metrics, indicating that distilling too many upper layers from the teacher can overly constrain the student representations. A similar trend is observed for Qwen3-Embedding 4B → BERT-base. The best results are obtained with 2–4 distilled layers, where Avg-Out reaches 81.44 and 81.39, respectively. Beyond this range, deeper distillation (6–12 layers) consistently reduces out-of-domain performance, despite relatively stable in-domain scores. This suggests diminishing returns when extending teacher supervision too deep into the student network.
These results indicate that $\mathcal{L}_\text{TAMD}$ is most effective when applied to a small number of top layers, where it provides strong semantic guidance from the teacher while preserving sufficient flexibility in lower layers. Excessive top-layer distillation, however, appears to limit the student’s ability to generalize, leading to reduced out-of-domain performance.

\paragraph{Ablation on Self Distillation.}
\begin{table*}[htbp]
\centering
\resizebox{\textwidth}{!}{%
\begin{tabular}{l|ccc|ccc|ccc|cc|c}
\toprule
& \multicolumn{3}{c|}{\textbf{Classification (F1)}} 
& \multicolumn{3}{c|}{\textbf{Pair Classification (AP)}}
& \multicolumn{3}{c|}{\textbf{STS (Spearman)}}
& \multicolumn{2}{c|}{\textbf{Domain Avg}} \\
\cmidrule(lr){2-10}\cmidrule(lr){11-12}
\textbf{Num Layers} & \textbf{Banking77} & \textbf{Tweet} & \textbf{Emotion$^{\star}$} 
& \textbf{MRPC} & \textbf{SciTail} & \textbf{WiC$^{\star}$} 
& \textbf{SICK} & \textbf{STS12} & \textbf{STSB$^{\star}$} 
& \textbf{Avg-In} & \textbf{Avg-Out} & \textbf{Avg.} \\
\midrule
\multicolumn{13}{c}{\textbf{Qwen3-Embedding 0.6B $\rightarrow$ MiniLMv2 H384}} \\
\midrule
2            
& 87.93 & 72.30 & 61.49 
& 83.39 & 80.27 & 68.76 
& 72.39 & 64.90 & 73.35 
& 67.87 & 76.86 & 73.86 \\
3
& 87.34 & 72.57 & 60.66 
& 85.23 & 81.30 & 67.43 
& 71.77 & 69.61 & 75.66 
& 67.92 & 77.97 & 74.62 \\
6     
& 86.69 & 72.77 & 60.94 
& 85.10 & 81.42 & 66.65 
& 72.51 & 70.77 & 76.22 
& 67.94 & 78.21 & 74.79 \\
\midrule
\multicolumn{13}{c}{\textbf{Qwen3-Embedding 4B $\rightarrow$ Bert-base}} \\
\midrule
2            
& 91.36 & 73.56 & 68.21 
& 84.97 & 81.29 & 70.06 
& 77.33 & 70.60 & 78.28 
& 72.18 & 79.85 & 77.30 \\
4
& 91.59 & 74.50 & 70.96 
& 86.30 & 81.42 & 69.63 
& 77.76 & 74.04 & 79.40 
& 73.33 & 80.94 & 78.40 \\
6     
& 91.66 & 74.56 & 70.92 
& 86.26 & 82.41 & 70.01 
& 78.08 & 74.81 & 80.35 
& 73.76 & 81.30 & 78.78 \\
9        
& 90.63 & 74.69 & 69.41 
& 86.52 & 82.40 & 68.48 
& 78.27 & 74.89 & 80.79 
& 72.89 & 81.23 & 78.45 \\
12        
& 91.43 & 74.30 & 70.71 
& 86.47 & 82.66 & 69.24 
& 78.38 & 75.40 & 80.88 
& 73.61 & 81.44 & 78.83 \\

\bottomrule
\end{tabular}
}
\caption{Ablation on $\mathcal{L}_\text{LASD}$ }

\label{tab:ablation_lasd}
\end{table*}

Table~\ref{tab:ablation_lasd} shows that increasing the number of self-distillation layers consistently improves performance across both teacher–student settings. For Qwen3-Embedding 0.6B → MiniLMv2 H384, out-of-domain performance rises steadily as more layers are included, reaching the best Avg-Out score at 6 layers. This indicates that deeper self-distillation strengthens representation robustness for smaller students. The effect is more pronounced for Qwen3-Embedding 4B $\rightarrow$ BERT-base, where performance improves monotonically from 2 to 12 layers, achieving the highest Avg-Out (81.44) and overall average (78.83) at the deepest configuration. Unlike $\mathcal{L}_\text{TAMD}$, no degradation is observed as depth increases, suggesting that larger students benefit from deeper self-distillation. Overall, these results show that $\mathcal{L}_\text{LASD}$  scales well with depth and consistently enhances generalization.

\paragraph{Performance of SAM Variants.}

\begin{table*}[htbp]
\centering
\resizebox{\textwidth}{!}{%
\begin{tabular}{l|ccc|ccc|ccc|cc|c}
\toprule
& \multicolumn{3}{c|}{\textbf{Classification (F1)}} 
& \multicolumn{3}{c|}{\textbf{Pair Classification (AP)}}
& \multicolumn{3}{c|}{\textbf{STS (Spearman)}}
& \multicolumn{2}{c|}{\textbf{Domain Avg}} \\
\cmidrule(lr){2-10}\cmidrule(lr){11-12}
\textbf{Method} & \textbf{Banking77} & \textbf{Tweet} & \textbf{Emotion$^{\star}$} 
& \textbf{MRPC} & \textbf{SciTail} & \textbf{WiC$^{\star}$} 
& \textbf{SICK} & \textbf{STS12} & \textbf{STSB$^{\star}$} 
& \textbf{Avg-In} & \textbf{Avg-Out} & \textbf{Avg.} \\
\midrule
\multicolumn{13}{c}{\textbf{Qwen3-Embedding 0.6B $\rightarrow$ MiniLMv2 H384}} \\
\midrule
SAM            
& 87.57 & 72.77 & 60.64 
& 83.54 & 80.49 & 68.04 
& 72.30 & 66.08 & 73.67 
& 67.45 & 77.13 & 73.90 \\
ASAM            
& 86.69 & 72.77 & 60.94 
& 85.10 & 81.42 & 66.65 
& 72.51 & 70.77 & 76.22 
& 67.94 & 78.21 & 74.79 \\
DISAM
& 86.84 & 73.11 & 60.52 
& 84.99 & 81.11 & 65.20 
& 72.30 & 69.80 & 75.43 
& 67.05 & 78.03 & 74.37 \\
\midrule
\multicolumn{13}{c}{\textbf{Qwen3-Embedding 4B $\rightarrow$ Bert-base}} \\
\midrule
SAM            
& 91.16 & 73.97 & 68.39 
& 85.10 & 80.84 & 70.38 
& 77.48 & 78.77 & 75.65 
& 71.47 & 81.22 & 77.97 \\
ASAM            
& 91.43 & 74.30 & 70.71 
& 86.47 & 82.66 & 69.24 
& 78.38 & 75.40 & 80.88 
& 73.61 & 81.44 & 78.83 \\
DISAM
& 91.27 & 74.60 & 71.14 
& 86.51 & 81.92 & 68.53 
& 78.57 & 76.00 & 80.53 
& 73.40 & 81.48 & 78.79 \\

\bottomrule
\end{tabular}
}
\caption{Comparison of SAM variants: SAM, DISAM, and ASAM }

\label{tab:ablation_sam}
\end{table*}

Table \ref{tab:ablation_sam} presents a comprehensive performance breakdown of SAM, DISAM, and ASAM across diverse benchmarks, including single-sentence classification (e.g., Banking77, Emotion), pair classification (e.g., MRPC, WiC), and semantic textual similarity (e.g., STS-B, SICK). Validated across two distinct teacher-student setups (Qwen3-Embedding 0.6B $\to$ MiniLMv2 H384 and Qwen3-Embedding 4B $\to$ Bert-base), the results demonstrate that ASAM consistently achieves the highest aggregated performance, recording average scores of \textbf{74.79} and \textbf{78.83}, respectively. While DISAM remains highly competitive—marginally outperforming ASAM in specific tasks such as Tweet sentiment analysis—ASAM exhibits superior overall robustness and stability across the full spectrum of evaluation metrics, justifying its adoption as the primary optimizer for the TALAS framework.

\end{document}